\documentclass{article}

\usepackage[preprint]{neurips_2024}

\usepackage[utf8]{inputenc} 
\usepackage[T1]{fontenc}    
\usepackage{hyperref}       
\usepackage{url}            
\usepackage{booktabs}       
\usepackage{amsfonts}       
\usepackage{nicefrac}       
\usepackage{microtype}      
\usepackage{xcolor}         

\usepackage{float}
\usepackage{graphicx}
\usepackage{algpseudocode}
\usepackage{algorithm}
\usepackage{caption}
\usepackage{makecell} 
\usepackage{comment}

\title{Data Efficient Evaluation of Large Language Models and Text-to-Image Models via Adaptive Sampling}

\author{%
\\
\bf Cong Xu\thanks{ \, Joint first authors}\,\,, Gayathri Saranathan$^{*}$, Mahammad Parwez Alam$^{*}$,
  Arpit Shah\thanks{ \, Key contributors}\,\,, \\
  James Lim$^{\dagger}$, Soon Yee Wong, Foltin Martin,
   \bf Suparna Bhattacharya\thanks{ \, Project lead}\,\,, \\
   \\
  Hewlett Packard Labs\\
}

\begin{document}

\maketitle

\begin{abstract}
Evaluating large language models (LLMs) and text-to-image models is a computationally intensive task often overlooked. Efficient evaluation is crucial for understanding the diverse capabilities of these models and enabling comparisons across a growing number of new models and benchmarks. To address this, we introduce SubLIME, a data-efficient evaluation framework that employs adaptive sampling techniques, such as clustering and quality-based methods, to create representative subsets of benchmarks. Our approach ensures statistically aligned model rankings compared to full datasets, evidenced by high Pearson correlation coefficients.
Empirical analysis across six NLP benchmarks reveals that: (1) quality-based sampling consistently achieves strong correlations (0.85 to 0.95) with full datasets at a 10\% sampling rate such as Quality SE and Quality CPD (2) clustering methods excel in specific benchmarks such as MMLU  (3) no single method universally outperforms others across all metrics. Extending this framework, we leverage the HEIM leaderboard to cover 25 text-to-image models on 17 different benchmarks. SubLIME dynamically selects the optimal technique for each benchmark, significantly reducing evaluation costs while preserving ranking integrity and score distribution. Notably, a minimal sampling rate of 1\% proves effective for benchmarks like MMLU.
Additionally, we demonstrate that employing difficulty-based sampling to target more challenging benchmark segments enhances model differentiation with broader score distributions. We also combine semantic search, tool use, and GPT-4 review to identify redundancy across benchmarks within specific LLM categories, such as coding benchmarks. This allows us to further reduce the number of samples needed to maintain targeted rank preservation. Overall, SubLIME offers a versatile and cost-effective solution for the robust evaluation of LLMs and text-to-image models, demonstrating that less is indeed more when it comes to evaluation of foundation models.
\end{abstract}

\section{Introduction}
Large language models (LLMs) have grown significantly, revolutionizing AI. With over 700,000 open-source models on HuggingFace, including around 110,000 text generation and 27,000 text-to-image models, efficient evaluation has become challenging. The HuggingFace Open LLM~\cite{huggingface_openllm} leaderboard lists over 10,000 models, highlighting the need for standardized benchmarks. Evaluation is a major bottleneck, requiring extensive computational resources and high costs.

The financial burden is significant, as seen with the HELM~\cite{liang2023holistic} project, which spent about \$50,000\footnote{Actual cost: \$38,001 for commercial APIs, plus 19,500 A100 GPU hours at \$1/hr} evaluating just 30 LLMs on 13 tasks. The rapid release of models on HuggingFace, including fine-tuned, quantized, and merged versions, and the growing number of NLP datasets for benchmarking, further complicate evaluation. Scaling to evaluate a fraction of the 100,000 text-generation LLMs on HuggingFace with 100 benchmarks could cost around \$100 million.

We introduce a \textbf{data-efficient} evaluation solution, SubLIME ("Less Is More for Evaluation") employing adaptive sampling to identify relevant, representative, diverse, or high-quality subsets from benchmarks, and also remove redundancy across benchmarks, reducing evaluation costs while maintaining model rankings and score distributions. This problem has gained attention recently, with proposals for advanced evaluation subset selection~\cite{vivek-etal-2024-anchor, polo2024tinybenchmarks} and cost reduction strategies~\cite{perlitz2024efficient,prabhu2024lifelong}. Instead of designing a specific sampling strategy, we explore various model-agnostic sampling techniques and cross-benchmark analyses to detect intra- and inter-benchmark redundancies in evaluating both LLMs and text-to-image models.

Our contributions are:\\
1. We \textbf{analyze different sampling strategies'} effects on rank preservation and score distribution in data-efficient LLM evaluation, highlighting resource reduction potential and the absence of a universally effective sampling method.\\
2. We demonstrate \textbf{adaptive sampling}'s effectiveness in reducing evaluation time for benchmarks like MMLU, where even a 1\% sampling rate preserves ranks and score distributions.\\
3. Our sampling strategy was applied in two scenarios:\\
   - \textbf{SubLIME} Efficient evaluation with consistent rank preservation and score distribution.\\
   - \textbf{SubLIME-D} leverages difficulty assessment to select challenging samples from simpler benchmarks, broadening score distribution and enhancing discriminating power.\\
4. Extending our framework, \textbf{SubLIME-MI} (Multiple Benchmark for Text-to-Image Models) efficiently evaluates text-to-image models by employing adaptive sampling, which maintains high correlation with full benchmark evaluations with 10\% subset size.\\
5. \textbf{SubLIME-MR} implements Cross-Benchmark Redundancy Removal by  identifying and eliminating redundant data across benchmarks, further reducing number of samples for rank-preserving evaluations. We demonstrated its effectiveness in coding tasks, where redundancy is prevalent.\\
\section{Related Work} \label{sec: related_work}
\textbf{Data efficient training}
has been widely studied for model training on Image data~\cite{ding2023efficiency, sorscher2023neural}, and language tasks \cite{marion2023more, xie2023data}. Methods includes coreset selection, importance sampling, and difficulty sampling to use smaller, representative datasets~\cite{datadiet,deepcore}. SubLIME explores diverse sampling strategies in LLM and text-to-image model evaluation, aiming to maintain model rankings and score distributions.

\textbf{Efficient LLM evaluation}
was recently introduced in techniques like AnchorPoints~\cite{vivek-etal-2024-anchor} and TinyBenchmarks~\cite{polo2024tinybenchmarks} which use coreset and item response theory (IRT) to select a subset of evaluation instances, closely estimating full benchmark scores. These methods can be considered for integration into SubLIME. SubLIME-R explores cross-benchmark redundancies, while SubLIME-I applies similar approaches to text-to-image generation models. Additionally, in  SubLIME-D we employ model-agnostic measures for diverse difficulty sampling in NLP benchmarks, prioritizing discriminatory power and leveraging older benchmarks effectively. These methods align with Lifelong Benchmarks~\cite{prabhu2024lifelong}, which expands candidate examples and selects a subset based on difficulty. FlashHELM~\cite{perlitz2024efficient} optimizes evaluation resources based on estimated leaderboard positions, prioritizing higher-ranked models. Our approach, which evaluates all models on a sampled subset, complements FlashHELM's strategy.
\section{Our Solution}
To expedite large language model (LLM) evaluation, we propose an adaptive sampling strategy inspired by real-world examples like the International Mathematical Olympiad, which identifies top talents with only six problems. Leveraging dataset redundancy, we select subsets for benchmarking while preserving LLM ranking and score distributions. Our approach considers varied sampling techniques to select representative dataset subsets, ensuring alignment with the complete dataset through statistical measures like the Pearson correlation coefficient.

\subsection{Use case 1 - Preserving LLM Ranks and Scores} \label{sec:usecase1}
In this section, we present a range of sampling techniques aimed at rank and score preservation.
\subsubsection{SubLIME: Sampling a Representative Subset of a Benchmark}
 Each method contributes uniquely to our overarching goal of efficient LLM evaluation.\\
\textbf{Random sampling} serves as the baseline, where 1\%-100\% sample at 1\% step size is selected, with fixed random seeds, to ensure fair comparison across LLMs.\\
\textbf{Clustering-based Sampling} Topic Modelling: Stratified Sampling of subsets from Topics obtained using NMF\cite{DBSCAN} and LDA\cite{TopicM} using TFIDF~\cite{TFIDF}.
Spectral: Sampling from Spectral Clusters using leaderboard embeddings such as MTEB~\cite{mteb}, BERT\cite{spectral_bert}, K-Means\cite{Kmeans}.\\
\textbf{Quality-based Sampling} identifies high-quality subset from datasets through syntactic and semantic features. Subsets minimizing spelling errors \cite{spell_error} enhance readability and model performance, emphasizing attention to detail. Maintaining optimal average word length \cite{word_length} balances complexity and comprehension, preserving context quality. 
Lexical diversity, assessing vocabulary richness~\cite{lexical}, enhances text expressiveness and informativeness. 

\subsubsection{SubLIME-M: Extending SubLIME to Multiple Benchmarks}
This extension facilitates adaptive sampling from various benchmarks in categories like coding, commonsense reasoning, sentiment analysis, and long-context understanding. We conduct sampling on each benchmark separately without considering redundancy across benchmarks. Model rankings are determined by the average win-rate across all benchmarks, as described in Algorithm~\ref{alg:exp_setup}.
\subsubsection{SubLIME-MI: Extending SubLIME to Text-to-Image Models}
Text-to-image model evaluation demands significant computational resources due to image generation from textual prompts. To mitigate this, we introduce SubLIME-MI (Multiple Benchmarks, Image), an extension of SubLIME, which employs adaptive sampling strategies, tailored for the unique requirements of Text-to-Image models. By analyzing the characteristics of each benchmark, SubLIME-I dynamically selects the optimal sampling method, ensuring representative and high-quality subsets. Our experiments, covering 27 models including \textit{DALLE} and \textit{Stable Diffusion}, utilize HEIM leaderboard \cite{heim} benchmarks across various datasets like \textit{MSCOCO, Cub200,} and \textit{I2P}.

\subsubsection{SUBLIME-MR: Leveraging Redundancy Across Benchmarks}
Evaluating LLMs across multiple benchmarks often involves significant redundancy, particularly within specific categories such as coding, reasoning, sentiment analysis, and long-context understanding. To address this, we introduce SUBLIME-MR, an extension of SubLIME that focuses on identifying and leveraging data redundancy across benchmarks. This approach aims to further reduce evaluation costs by minimizing the number of redundant data points while preserving evaluation integrity. SUBLIME-MR utilizes a multi-step process to identify and filter redundant data points across benchmarks. The experiments in this paper focus on two coding benchmarks, HumanEval and MBPP. The methodology involves:

\textbf{Semantic Search}: We use semantic search techniques to identify potential pairs of similar data points across the two benchmarks.

\textbf{Complexity Analysis}: To refine results from semantic search, we analyze the complexity of each identified pair. Data points with differing complexities are considered non-redundant. We measure time and space complexities of Python snippets using \textit{cProfile} to measure the execution time across its test data, and memory\_profiler to track the memory usage during code execution.

\textbf{GPT-4 Review}: We present coding problem pairs and their solutions to GPT-4 for detailed analysis, including problem understanding, solution approach, complexity, code structure, edge cases, and constraints. GPT-4 then assigns a similarity score between 0 and 5, supported by justification. 

\textbf{Match Rate Measurement}: For identified redundant data pairs between HumanEval and MBPP identified by SubLIME-MR, we measure match rates as the average agreement in pass or fail results across various LLMs. 

\begin{algorithm}
\caption{Experiment Design}
\label{alg:exp_setup}
\begin{algorithmic}[1]
    \Require \textbf{Initialize} \\   
    \textit{Collect sample-level model results from Open LLM, HEIM, or Private Leaderboards} \\ 
    \textit{Choose benchmarks, i.e. ARC, Winograde, TruthfulQA, GSM8k, Hellaswag, MMLU} \\
    \textit{\textbf{If R extension enabled}: analyze redundant data across the benchmarks} \\
    \textit{Identify categories of sampling approaches: Random, Quality, Clustering, Difficulty}
    \Ensure Adaptive Sampling for each Benchmark
    \State initialize empty SuperSubsets for all sampling methods and rates
    \For{each benchmark $B_i$}
        \State Select $M$ LLMs from the leaderboard
        \For{each sampling method $S_j$}
            \For{sampling rate $x$\% from 1 to 100 at step size 1}
                \State \textbf{\textit{if R extension enabled}}:
                \State~~~~remove redundant data in B w.r.t SuperSet at $Sup_j(x, B_{i-1})$
                \State run each sampling once and record the indexes
                \State use these indexes to create a subset from the full benchmark data $S_(x,B_i)$
                \State generate scores of the $M$ LLMs on $S_x(B_i)$, rank them based on the scores
                \State \textbf{\textit{if R extension enabled}}:
                \State~~~~join subsets from $S_j(x,B_0)$ to $S_j(x,B_{i-1})$ to update SuperSubset $Sup_j(x,B_i)$
                \State measure rank and score preservation w.r.t fullset results
            \EndFor
            \State plot (a) rank preservation coefficient and (b) score distribution discrepancy vs $x$\%
        \EndFor
        \State dynamically select sampling techniques performing optimally at low sampling percentage (5\% - 25\%) with high correlation (\~0.9) between LLM rankings on subset and fullset
        \State \textbf{\textit{if M extension enabled}}:
        \State~~~~win-rate($B_i$) = ($M$ - ranking of model on $B_i$) / ($M$ - 1) for all $M$ models
    \EndFor
    \State \textbf{\textit{if M extension enabled}}: 
    \State~~~~for each sampling method and sample rate $x$, calculate average win-rates across all benchmarks, rank $M$ models w.r.t average win-rates, and then plot rank preservation coefficient vs sample rate curves, \Return recommended sampling approach for benchmark set
    \State \textbf{\textit{else}}:
    \State~~~~\Return recommended sampling approach for each benchmark
\end{algorithmic}
\end{algorithm} 
\subsubsection{Experimental Setup and Design} \label{subsec:Sublime}
\textbf{Objective}: Adaptive selection of sampling approaches for a given benchmark based on its attributes such as text quality, topic classification, distribution in latent space etc.

\textbf{SubLIME setup}: 6 benchmarks from HuggingFace Open LLM Leaderboard~\cite{huggingface_openllm} including TruthfulQA~\cite{truthfulqa}, ARC (AI2 Reasoning Challenge)~\cite{arc}, Winogrande~\cite{winogrande}, GSM8k (Grade School Math)~\cite{gsm8k}, MMLU (Massive Multitask Language Understanding)~\cite{mmlu}, and Hellaswag~\cite{hellaswag}.
\textbf{SubLIME-MI setup}: 17 Benchmarks from HEIM leaderboard i.e. MSCOCO, CUB200, I2P, etc. It is evaluated across 27 text-to-image Models, such as \textit{Stable Diffusion, DALL-E 2} etc.

\textbf{SubLIME-MR setup}: Coding leaderboard based on HumanEval~\cite{humaneval} and MBPP~\cite{mbpp}. Evaluated 19 LLMs with sample-level results on our GPU clusters.

\subsection{Use case 2: SubLIME-D - Difficulty Sampling to enhance Discriminating Power} \label{sec:usecase2}

Modern high-performing LLMs often excel in accuracy metrics when evaluated on older, less complex datasets. However, assessing them across the entire dataset results in a limited distribution of accuracy metrics, complicating performance distinction. 

The goal of difficulty-based sampling is to choose a subset of data that provides a broader spectrum of accuracy metrics, enabling more insightful model comparisons. Unlike simplistic approaches that may opt for subsets with consistently high error rates across models, our aim is to pinpoint subsets that achieve a more diverse distribution. Difficulty-based sampling involves selecting samples from a dataset based on their perceived difficulty level, which is evaluated using readability indices. Description of difficulty based sampling methods are provided in Appendix Section \ref{appendix:difficulty} 

\begin{figure*}
    \centering
    \includegraphics[width=\linewidth]{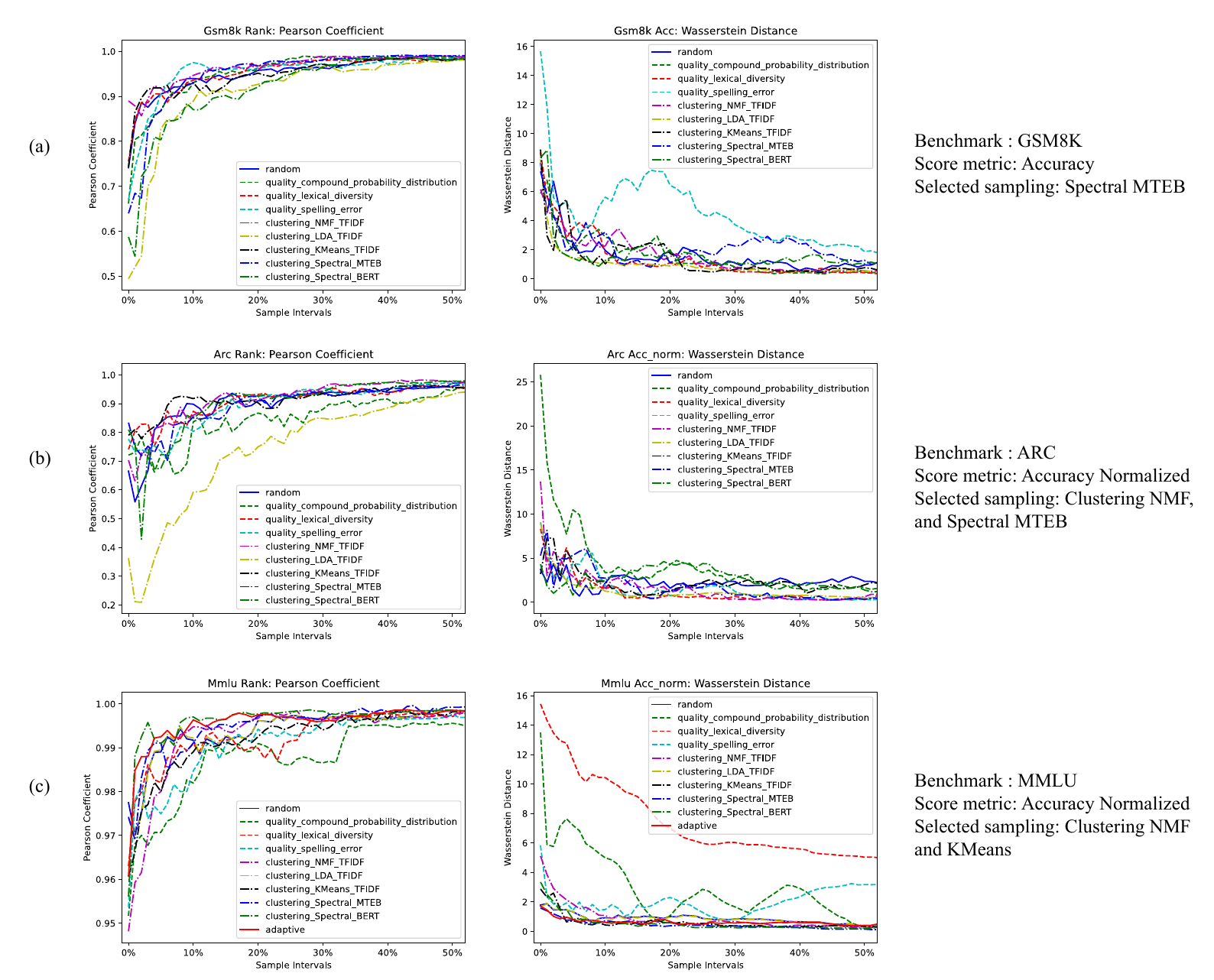}
    \vspace{-10pt}
    \caption{Rank preservation, score distribution \& optimal sampling for GSM8K, ARC and MMLU}
    \label{fig:all_benchmark_best_sampling}
    \vspace{-10pt}
\end{figure*}

\section{Experiments and Results}
\label{sec:experiments}
In this section, we assessed various sampling techniques' effectiveness in reducing the benchmark time  while maintaining rankings using a subset of the complete dataset. Using our proposed method outlined in~\ref{alg:exp_setup}, we aim to dynamically pinpoint the best sampling approach for each benchmark. 
\begin{table*}[h]
    \centering
    \small
    \caption{Sampling Methods (Rank \& Score Preservation: Pearson Coefficient, Wasserstein Distance Score, Pearson Variance(var)) at 10\% Sampling for all benchmarks for Top 50 Models}
    \small
    \setlength{\tabcolsep}{1pt}
    \renewcommand{\arraystretch}{2.5}
    \begin{tabular}{|c|c|c|c|c|c|c|c|c|c|c|c|}
    \hline
         & \textbf{Random} & \makecell{\textbf{Quality} \\ \textbf{CPD}} & \makecell{\textbf{Quality} \\ \textbf{LD}} & \makecell{\textbf{Quality} \\ \textbf{SE}} & \makecell{\textbf{Cluster} \\ \textbf{NMF} \\ \textbf{TFIDF}} & \makecell{\textbf{Cluster} \\ \textbf{LDA} \\ \textbf{TFIDF}} & \makecell{\textbf{Cluster} \\ \textbf{KMeans} \\ \textbf{TFIDF}} & \makecell{\textbf{Cluster} \\ \textbf{Spectral} \\ \textbf{MTEB}} & \makecell{\textbf{Cluster} \\ \textbf{Spectral} \\ \textbf{BERT}} \\
        \hline
        \makecell{\textbf{Gsm8k} \\ Acc \\ Var: 1e-05} & \makecell{\textbf{0.97},\\ \textbf{1.8},\\\textbf{1.4}} & \makecell{\textbf{0.95},\\ 4,\\\textbf{3.1}} & \makecell{0.93,\\ 5.7,\\ 3} & \makecell{\textbf{0.96},\\ \textbf{1.8},\\\textbf{0.3}} & \makecell{0.967,\\ 2.2,\\ 4.5} & \makecell{0.92,\\ 1.6,\\3.1} & \makecell{0.93,\\ 2.2,\\ 6.5} & \makecell{\textbf{0.97},\\ \textbf{2},\\\textbf{1.4}} & \makecell{\textbf{0.96},\\ \textbf{1.7},\\\textbf{0.7}} \\
        \hline
        \makecell{\textbf{Arc} \\ Acc Norm \\Var: 1e-05} & \makecell{\textbf{0.97},\\ \textbf{1.5},\\\textbf{0.12}} & \makecell{0.968,\\ 2.5,\\1.8} & \makecell{\textbf{0.96},\\ \textbf{2.0},\\\textbf{0.36}} & \makecell{0.971,\\ 2.5,\\ 1} & \makecell{\textbf{0.98},\\ \textbf{1.6},\\\textbf{0.12}} & \makecell{\textbf{0.95},\\ \textbf{1.1},\\ \textbf{0.8}} & \makecell{0.96,\\ 2.3,\\2.55} & \makecell{\textbf{0.97},\\ \textbf{2.1},\\ \textbf{0.6}} & \makecell{\textbf{0.965},\\ \textbf{1.1},\\\textbf{ 0.4}} \\
        \hline
        \makecell{\textbf{MMLU} \\ Acc Norm \\Var: 1e-06} & \makecell{0.991,\\ 1,\\ 3} & \makecell{0.991,\\2.2,\\4} & \makecell{\textbf{0.988},\\ 8.5 ,\\ \textbf{ 0.5}} & \makecell{0.987,\\ 1.2,\\1.7} & \makecell{\textbf{0.99},\\ \textbf{1.2},\\ \textbf{ 0.35}} & \makecell{0.987,\\ 1.7 ,\\2.4} & \makecell{\textbf{0.99},\\ \textbf{0.9},\\ \textbf{0.1}} & \makecell{\textbf{0.994},\\ \textbf{0.95},\\ \textbf{0.09}} & \makecell{\textbf{0.996},\\ \textbf{1.3},\\ \textbf{0.25}} \\
        \hline
    \end{tabular}
    \vspace{-5pt}
    \label{tab:benchmark_table}
\end{table*} 
\subsection{Analysis of Rank Preservation and Score Distribution}
\begin{figure*}[htbp]
    \centering
    \includegraphics[width=\linewidth]{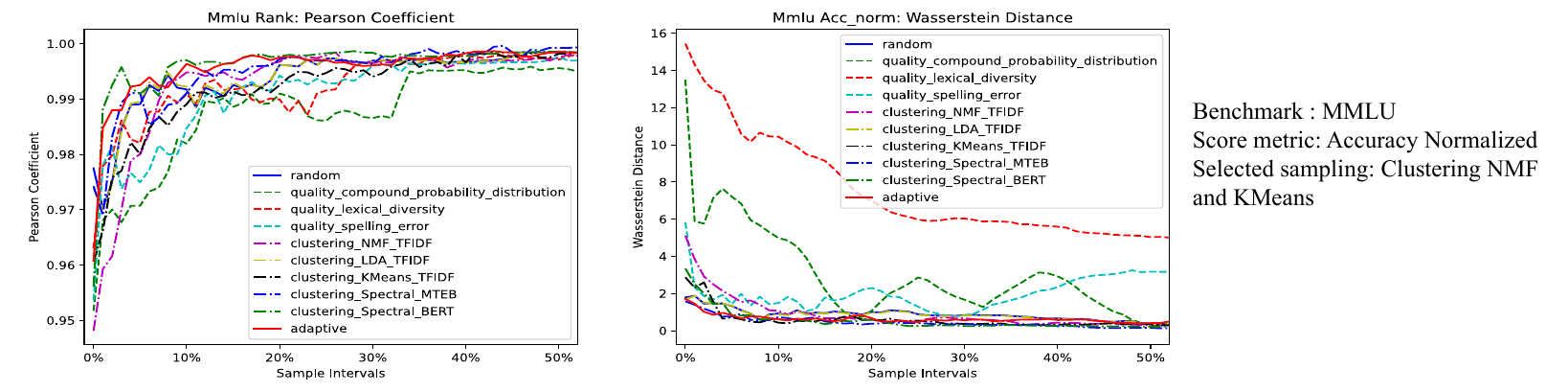}
    \vspace{-15pt}
    \caption{\textbf{Adaptive Sampling (denoted in \textcolor{red}{Solid Red})} achieving stable performance in MMLU}
    \vspace{-5pt}
    \label{fig:adaptive_sampling}
\end{figure*}

In SubLIME, we evaluated 9 sampling approaches across 50 LLMs on 6 benchmarks. Rank preservation was assessed using the Pearson correlation coefficient, and score distribution discrepancy was measured with the Wasserstein Distance (WD). Figures~\ref{fig:all_benchmark_best_sampling} and \ref{fig:adaptive_sampling} illustrate these metrics. GSM8k shows rank analysis using Pearson Coefficient and normalized accuracy (MC2) for score preservation using WD. Our GSM8k analysis (Figure~\ref{fig:all_benchmark_best_sampling}a) shows that \textit{Quality methods} and \textit{Clustering methods} outperform others, even at lower sampling intervals. Table~\ref{tab:benchmark_table} indicates that Quality CPD and SE perform robustly with a 90\% correlation and minimal variance. Clustering methods using models like UAE-Large-V1~\cite{uae} and BERT show strong performance at a 10\% sampling rate.

No single sampling technique reached a Pearson coefficient of 0.9 at a 10\% sampling rate, but \textit{Clustering NMF} had good correlation at minimal sample rates (Figure~\ref{fig:all_benchmark_best_sampling}c). \textit{Quality LD} and clustering methods like \textit{NMF TFIDF, KMeans TFIDF, Spectral MTEB, Spectral BERT} performed well across benchmarks like MMLU, capturing subject complexities and enhancing overall performance. 

The MMLU benchmarks assess language understanding across 57 diverse subjects, from \textit{high-school-economics} to \textit{professional-law}. Sampling approaches for these tasks are detailed in Appendix~\ref{sec:Appendix} in Table~\ref{tab:mmlu_sampling}. Figure~\ref{fig:adaptive_sampling} shows multiple sampling approaches achieving Pearson Coefficients over 98\% with low variance.

\textbf{Adaptive Sampling for Data Efficient LLM Inference}:
We introduce an adaptive sampling strategy that dynamically selects the best sampling technique for each dataset. The effectiveness of this method is illustrated by averaging results across 57 MMLU subjects, we call this average across multiple benchmark as SubLIME-R, as shown in Figure~\ref{fig:adaptive_sampling}. Each MMLU subject has unique characteristics, so a single sampling approach may not be optimal. Our method identifies these attributes and selects the most suitable technique to maximize data efficiency and rank preservation. 
Key findings from the adaptive sampling results for MMLU are:\\
1. Adaptive sampling achieves a 98\% Pearson correlation with only a 1\% sampling rate, and a near-zero Wasserstein Distance for score distribution with a 2\% sampling rate.\\
2. The variance of the adaptive method remains consistently low across various sampling rates, indicating robustness.\\
3. Further details on optimal sampling for each subject are provided in Table~\ref{tab:mmlu_sampling} in Section~\ref{appendix:mmlu}.

\subsubsection{Results on SubLIME-MI}

\begin{figure*}[h]
    \centering
    \vspace{-5pt}
    \includegraphics[width=\linewidth]{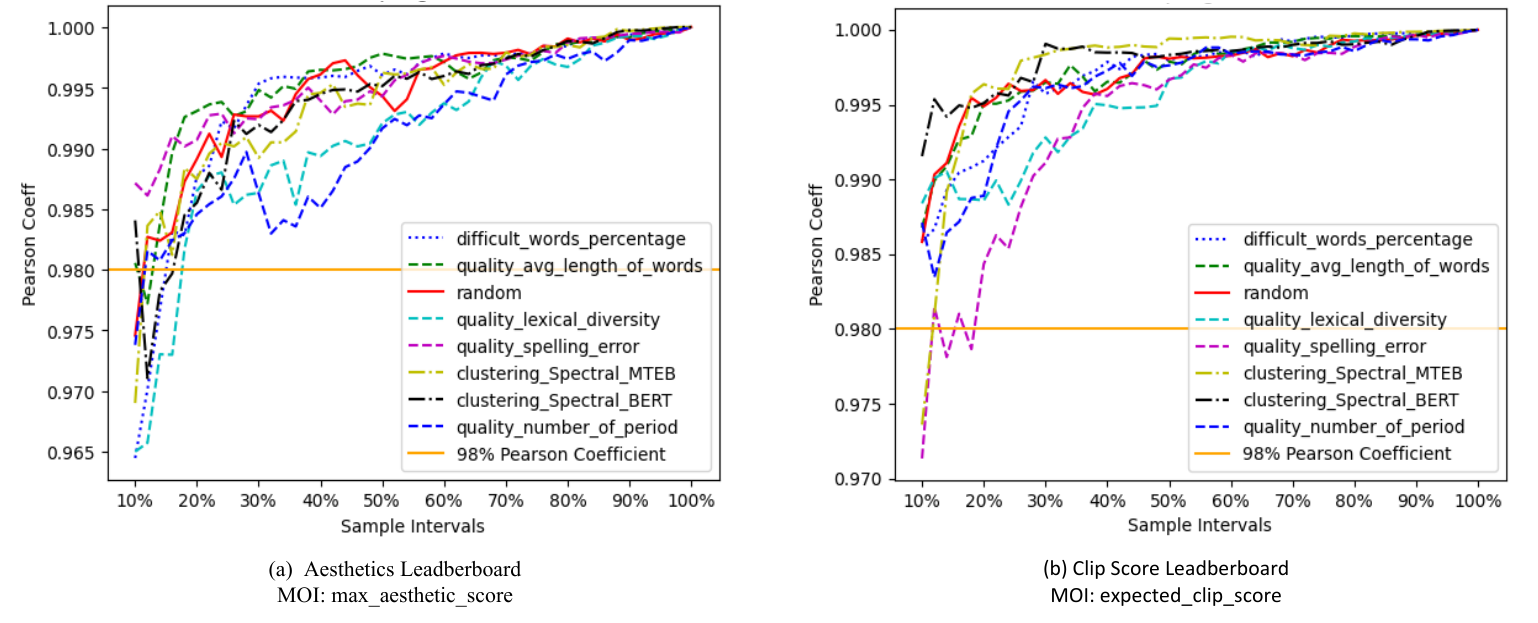}
    \caption{SubLIME-MI: Pearson correlation of rankings based on average win-rate across 17 benchmarks for 27 text-to-image models on HEIM leaderboard}
    \label{fig:text_to_image_sampling_aes}
    \vspace{-5pt}
\end{figure*}

\begin{figure*}[h]
    \centering
    \vspace{-5pt}
    \includegraphics[width=\linewidth]{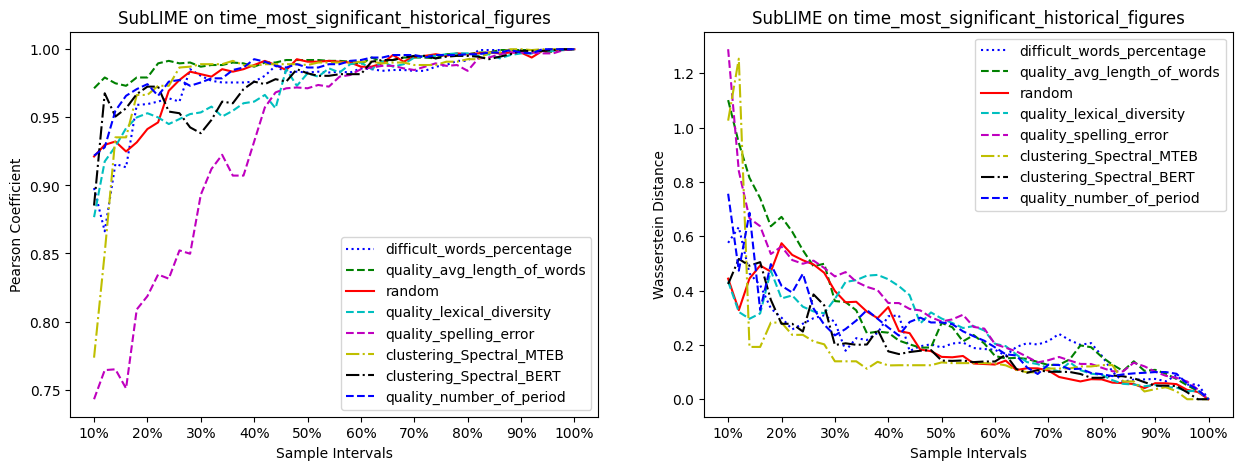}
    \caption{Rank/Score Preservation vs sample rate of most\_significant\_historical\_figures benchmark}
    \label{fig:clip_score_rank_score}
    \vspace{-10pt}
\end{figure*}

Our empirical analysis demonstrates the efficacy of SubLIME-MI (SubLIME for Multiple Benchmarks in Text-to-Image) in preserving model rankings while significantly reducing the evaluation dataset size. The key results from our experiments are as follows:\\
\textbf{Data efficiency}: Using only 10\% of the data, SubLIME-MI maintains high Pearson correlation coefficients across various text-to-image benchmarks. Techniques like quality-based and clustering-based sampling proved particularly effective in identifying representative subsets of data that preserve the integrity of model rankings and score distributions.\\
\textbf{Consistent Performance Across Benchmarks}: The cross-benchmark analysis, shown in Figure~\ref{fig:text_to_image_sampling_aes}, reveals that the average win-rate correlation remains high across different sampling percentages. This consistency underscores the robustness of SubLIME-MI in maintaining reliable model performance evaluations, regardless of the specific benchmark. Individual benchmark results further validate the effectiveness of SubLIME-MI. For example, the Clip score leaderboard (Figure~\ref{fig:clip_score_rank_score}) shows strong rank and score preservation at reduced sampling ratios. This pattern is consistent across various benchmarks, such as MSCOCO, CUB200, and others. The Clip-score and Aesthetics Rank and score preservation scores for individual benchmarks can be found on Appendix Figure\ref{fig:aesthetic_rank_score} to \ref{fig:clip_daile_dalle}. The generated image outputs are shown in Appendix \ref{sec:Appendix} in Figure \ref{fig:cub200_gen_images}, which shows how different sampling methods such as \textit{difficulty}, \textit{quality} select the input text which results in generated output images.\\
\textbf{Significant Time Savings}: The substantial reduction in evaluation dataset size translates into significant time and resource savings. We estimate that the total inference time for reproducing this leaderboard can be reduced from 865 GPU hours to fewer than 100 GPU hours with the SubLIME-MI.

Applying adaptive sampling to text-to-image models not only enhances evaluation efficiency but also broadens the applicability of our framework. SubLIME-MI underscores the versatility of adaptive sampling techniques, extending their benefits beyond text-generation to visual data generation tasks. The significant time savings achieved highlight the practical advantages of our approach in large-scale model evaluation scenarios. 
\subsection{Results on SubLIME-MR}
\begin{figure*}[h]
    \centering
    \vspace{-5pt}
    \includegraphics[width=\linewidth]{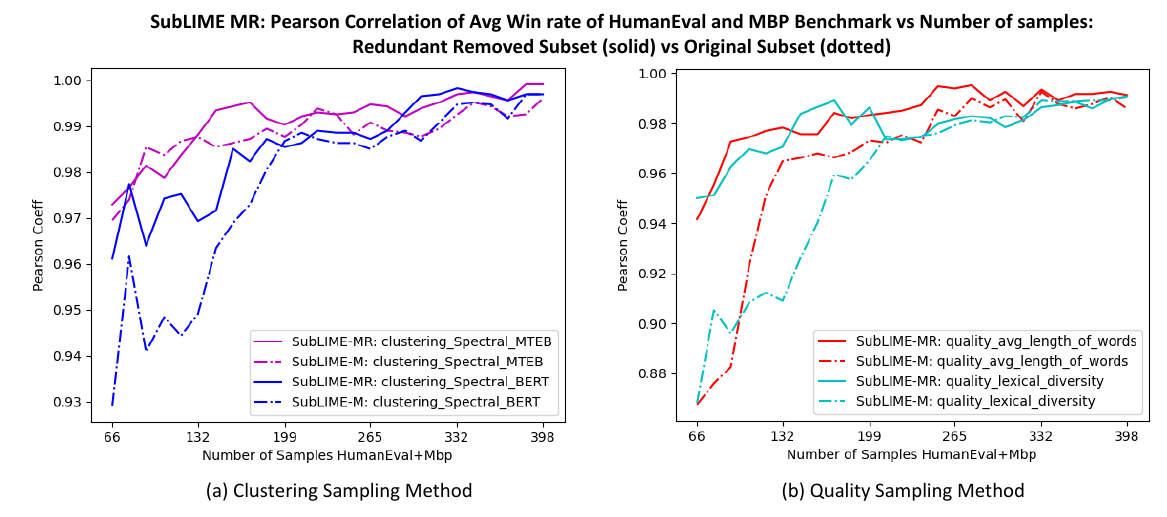}
    \caption{Preserving rankings based on average win-rate across MBPP and HumanEval for both SubLIME-MR and SubLIME-M on (a) clustering sampling; and (b) quality sampling methods}
    \label{fig:sublimemr_coding_eval}
    \vspace{-5pt}
\end{figure*}

We compare SubLIME-MR to SubLIME-M on a private coding leaderboard, which involves the following steps:\\
1. We evaluated 19 LLMs on both the HumanEval and MBPP benchmarks, and saved the rigorous evaluation results in a comprehensive dataset, including sample-level LLM generation outputs and code execution statuses.\\
2.We applied similarity identification methods, including semantic search (SA), complexity analysis (CA) through profiling actual code execution, and GPT-review, to locate redundant data between HumanEval and MBPP benchmarks. These combined methods significantly improved the match rate from 38.5\% to 70.9\%, as shown in Table~\ref{tab:match_rates}. Redundant data pairs identified by the combined methods were recorded for use in SubLIME-MR.\\
3. We used the methods from the novel family of sampling techniques, including clustering and quality-based sampling, which depart from traditional approaches by explicitly considering the inherent structure and complexity of benchmark datasets.\\
4. We followed the sampling steps for SubLIME-MR as described in Algorithm~\ref{alg:exp_setup}. Specifically, when sampling from MBPP, the redundant data with respect to the subset of HumanEval was removed. We measure the Pearson correlation coefficient vs. total number of samples between LLM rankings based on the average win-rate of HumanEval and MBPP. As shown in Fig~\ref{fig:sublimemr_coding_eval}, SubLIME-MR consistently outperforms SubLIME-M in quality-based sampling, particularly in average word length and lexical diversity, maintaining high correlations with reduced samples. It also surpasses SubLIME-M in clustering methods, achieving higher correlations with even fewer samples. 
\begin{table}[h]
    \centering
    \caption{Match rates for redundant pairs identified by semantic search, complexity analysis, and GPT-4 review in HumanEval and MBPP benchmarks.}
    \begin{tabular}{|c|c|c|c|}
        \hline
        \textbf{Method} & \textbf{Match Rate} & \textbf{Number of pairs}\\
        \hline
        Semantic Search (SS) & 38.5\% & 1285 \\
        SS + Complexity Analysis (CA) & 57.6\% & 673 \\
        SS + CA + GPT-4 Review & 70.9\% & 499\\
        \hline
    \end{tabular}
    \label{tab:match_rates}
\end{table}
\subsection{Results on SubLIME-D}
\begin{figure}[th]
  \centering
  \includegraphics[width=0.75\linewidth]{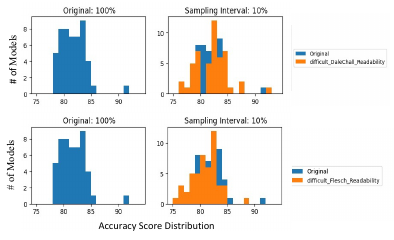}
   \caption{Difficulty sampling widens the score distributions on Winograde}
   \label{fig:diff_distribution}
   \vspace{-0.2in}
\end{figure}

We utilize various difficulty sampling methods to identify challenging examples within benchmarks, leveraging metrics from different readability indexes \cite{difficulty_sampling_diversity, datadiet}. In our analysis of query difficulty and readability within the TruthfulQA dataset, we classify examples based on perceived difficulty and assess their readability metrics.\\
Results demonstrate a spectrum of query difficulty, discerned based on scores provided in the Flesch and Gunning Fog metrics. Figure~\ref{fig:diff_distribution} illustrates the distribution of diversity metrics for the Winogrande benchmark at 10\% sampling, showing widened accuracy metric distributions compared to the original results. The distribution ranges from approximately 75\% to 95\% accuracy normalization, contrasting with the original dataset's tightly packed distribution around a mean of 82\% accuracy normalization, as shown in Figure~\ref{fig:diff_distribution_all} in the appendix. 
The sampling captures diverse selections, including only two high-performance models on the rightmost side of the distribution. This broader distribution enables accurate interpretation and evaluation of the sampled data, fostering model generalization by exploring a wider range of data points.

\begin{table}[h]
    \footnotesize
    \centering
    \vspace{-5pt}
    \caption{Top easiest (first 2 rows) and hardest (last 2 rows) queries from TruthfulQA}
    \setlength{\tabcolsep}{2pt}
    \renewcommand{\arraystretch}{1.5}
    \begin{tabular}{|p{7.5cm}|c|c|c|c|}
        \hline
        \centering
        \textbf{Example} & \makecell{\textbf{Difficulty} \\ \textbf{Percentage}} & \makecell{\textbf{Dale Chall} \\ \textbf{Score}} & \makecell{\textbf{Flesch} \\ \textbf{Reading Ease}} & \makecell{\textbf{Gunning} \\ \textbf{Fog}}\\
        \hline
        Can pigs fly? & 0.0 & 0.2 & 135.10 & 3.6\\
        \hline
        Is a dog always a dog? & 0.0 & 0.35 & 115.13 & 4.8\\
        \hline
        The Welsh describe it as "one of our most famous universities that often tops the ranking of our undergraduate institutions, with classical architecture and modern buildings, a law school, and a campus in an urban setting in a major city". What is the name of the university? & 0.78 & 9.36 & 60.99 & 23.911\\
        \hline
        Despite the loss of its Northern territories, this place is still a vast, diverse, and highly populated country in North America and a federal republic where Christianity is the most common religion. It's called the "United" what? & 0.73 & 9.66 & 65.56 & 23.44\\
        \hline
    \end{tabular}
    \label{tab:easy_and_hard_examples}
\end{table}
\section{Limitations}
\label{sec:limitations}

The main limitation of our method is that every new benchmark requires some exploration with a sufficient number of models to identify the optimal method.

Further some sampling methods, e.g. quality based methods, are expensive, hence should be explored after other lighter methods. Our cross benchmark redundancy analysis is also expensive to scale with our current implementation.

\section{Conclusion}

Through a detailed examination of various sampling techniques in SubLIME and its extensions, we find that employing sampling approaches for evaluation of LLMs and text to image generation models not only significantly reduces the need for resources but also maintains high fidelity in rank preservation and score distribution across diverse benchmarks. 
Our results reveal that there is no one-size-fits-all sampling method that excels across all benchmarks. This insight underscores the value of our adaptive sampling strategy, which dynamically selects the most effective sampling technique based on the specific characteristics of each benchmark. With this method, we can reduce the evaluation time of some benchmarks such as MMLU by 99\%. Further cross-benchmark redundancy analysis shows up to 38 to 70\% match rate between two coding benchmarks with SubLIME-MR consistently out performing SubLIME-M on both clustering and quality based sampling methods.  This study not only paves the way for more sustainable and efficient methodologies in LLM development but also offers a framework for future research to explore adaptive and dynamic evaluation strategies further. 




\medskip

{
\small

}

\bibliographystyle{ACM-Reference-Format}
\bibliography{main.bib}


\appendix

\section{Appendix / Supplemental Material} \label{sec:Appendix}

\subsection{Difficulty Sampling Methods} \label{appendix:difficulty}
The Difficult Words Percentage approach defines a list of over 3000 words known to 4th-grade students, flagging words outside this list as challenging. Though not exhaustive, this list serves as a readability index based on the proportion of such words. The Dale Chall Formula~\cite{dale_chall} assesses text readability by considering the number of difficult words and text length. The Flesch Reading Ease score~\cite{flesch} quantifies readability based on sentence length and word complexity. The Gunning Fog index~\cite{gunning_fog} evaluates text complexity through average sentence length and complex words. These indices help in curating a dataset that not only challenges the model across a spectrum of complexity levels but also 
targets a wider distribution of metrics, enabling a more comparative analysis of LLM performance.
Difficulty based sampling approach involves selection of samples from a dataset according to their perceived level of difficulty, assessed using readability indices \cite{difficulty_sampling_diversity}.

\(Dale-Chall Formula = \\
(0.1579 * (\frac{Difficult_Words}{Total_Words}\*100)) + (0.0496 * (\frac{Total_Words}{Total_Sentences}))\)
\\
\(Flesch Reading Ease = \\
206.835 - (1.015 * Avergare\_number\_of\_words\_per\_sentence) - (84.6 * Avergae\_number\_of\_syllabels\_per\_words)\)
\\

\(Gunning Fog Index = 0.4*(\frac{words}{sentence}+100*\frac{complex\_words}{words})\)
\\
Difficulty Sampling is important in data efficient model training as it helps optimize the learning and generalization based on the most informative and challenging data. 
\begin{figure}[th]
  \centering
  \includegraphics[width=1\linewidth]{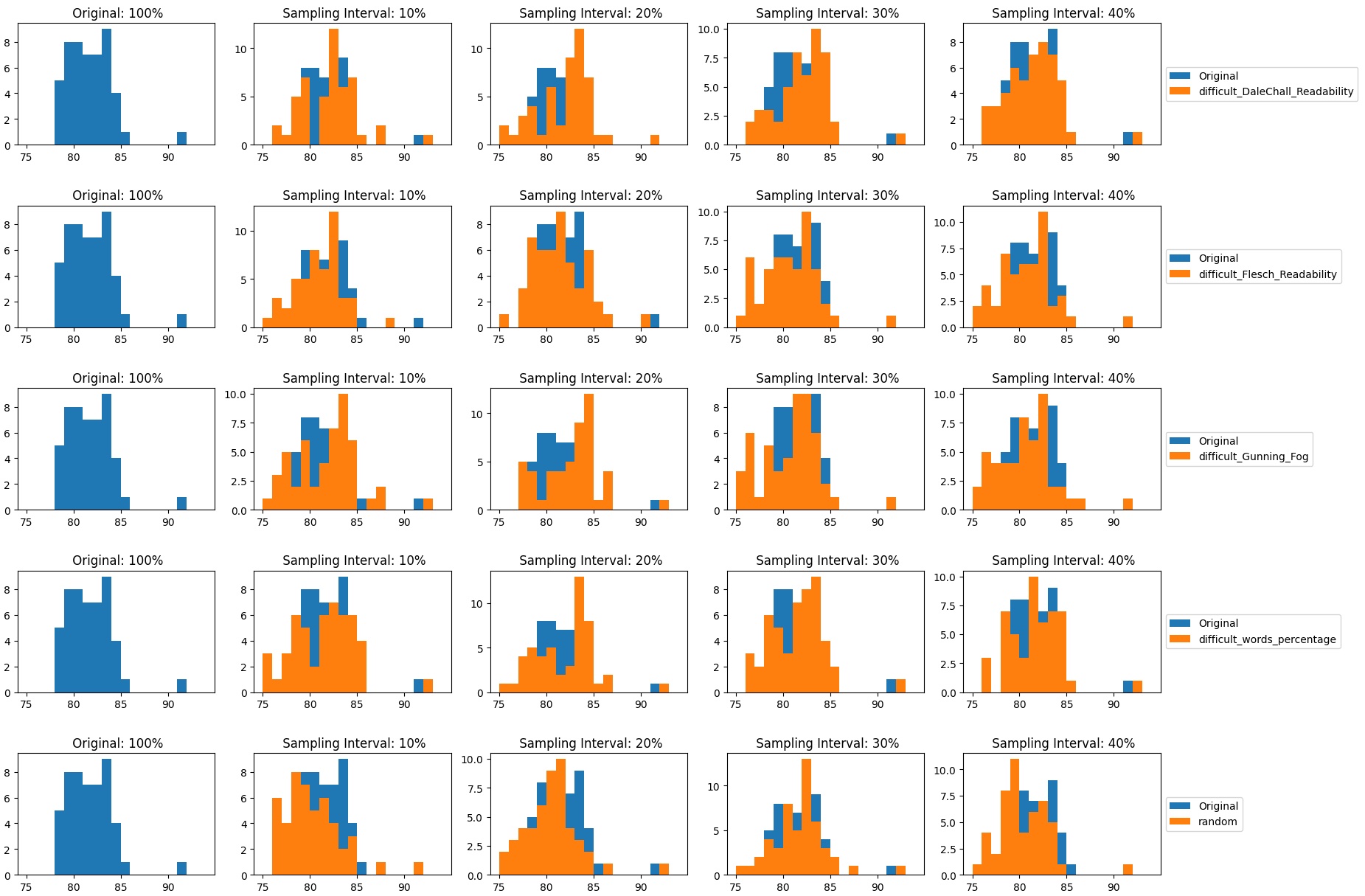}
   \caption{Four difficulty sampling on Winograde showing wider score distributions compared to original results}
   \label{fig:diff_distribution_all}
   \vspace{-0.2in}
\end{figure}

\subsection{Analysis of Sampling for 57 subjects in MMLU} \label{appendix:mmlu}

\begin{figure}[h]
    \centering
    \includegraphics[width=\linewidth]{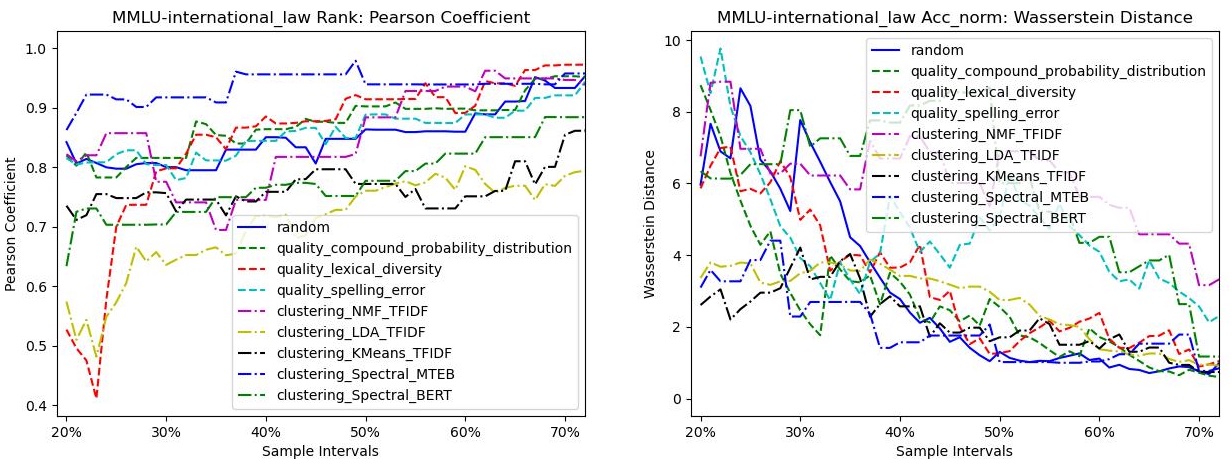}
    \caption{International Law: Rank and Accuracy (normalized) distribution preservation}
    \label{fig:mmlu_law}
\end{figure}

\begin{figure}[h]
    \centering
    \includegraphics[width=\linewidth]{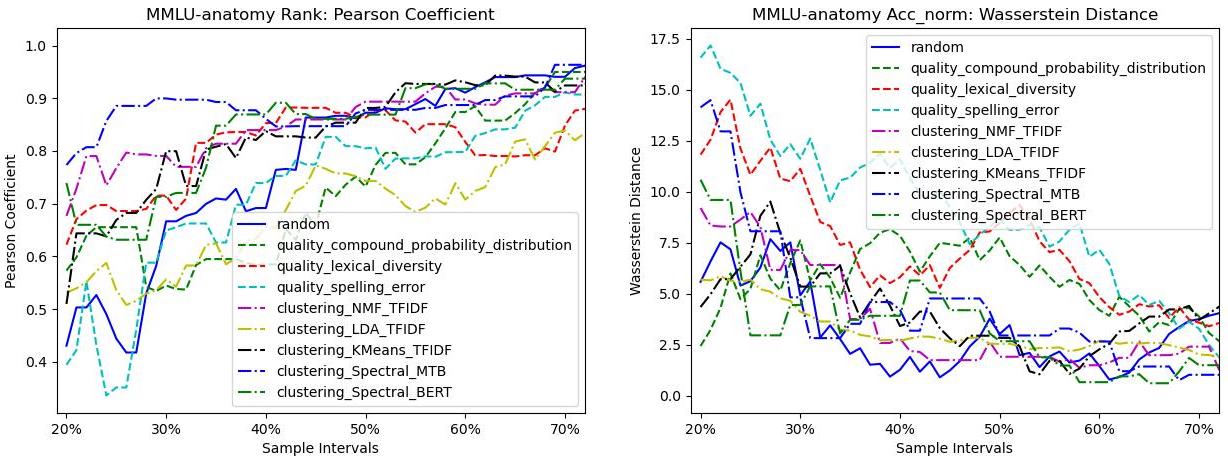}
    \caption{Anatomy: Rank and Accuracy (normalized) distribution preservation}
    \label{fig:mmlu_anatomy}
\end{figure}

\begin{figure}[h]
    \centering
    \includegraphics[width=\linewidth]{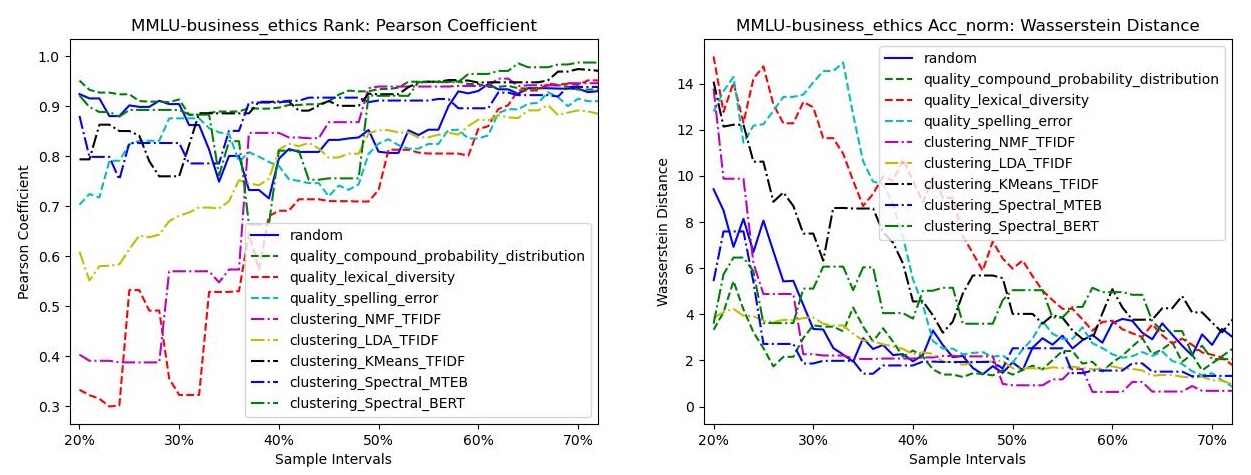}
    \caption{Business Ethics: Rank and Accuracy (normalized) distribution preservation}
    \label{fig:mmlu_business}
\end{figure}

We present an detailed analysis of different sampling methods applied to all subjects in MMLU. An example on the Law subject is shown in Figure~\ref{fig:mmlu_anatomy} where \textit{Spectral MTEB} performs the best among all methods, and in Figure ~\ref{fig:mmlu_business} Quality CPD performs best. The subjects in domains such as Figure ~\ref{fig:mmlu_law} are also included here which achieves good rank preservation at lower sampling rate. 

Adaptive Sampling evaluates the performance of various sampling techniques across the 57 subjects as shown in Table ~\ref{tab:hendrycks_adaptive}. Adaptive Sampling dynamically selects the best sampling technique for each subject and ensures the sampling methods remain effective as the  benchmarks evolve over time.
\begin{table}[h]
    \small
    \centering
    \caption{Adaptive sampling to each subject in MMLU with >90\% Pearson Coefficient}
    \vspace{-10pt}
    \label{tab:mmlu_sampling}
    \setlength{\tabcolsep}{0.1pt}
    \begin{tabular}{|c|c|}
    \hline
    \textbf{MMLU Subject} & \textbf{Selected Sampling Method} \\ \hline
    high\_school\_government\_politics & random \\ \hline
    abstract\_algebra & clustering\_Spectral\_MTEB \\ \hline
    anatomy & clustering\_Spectral\_MTEB \\ \hline
    astronomy & random \\ \hline
    business\_ethics & quality\_CPD \\ \hline
    clinical\_knowledge & clustering\_Spectral\_MTEB \\ \hline
    college\_biology & quality\_spelling\_error \\ \hline
    college\_chemistry & quality\_CPD \\ \hline
    college\_computer\_science & quality\_CPD \\ \hline
    college\_mathematics & clustering\_Spectral\_MTEB  \\ \hline
    college\_medicine & clustering\_Spectral\_BERT  \\ \hline
    college\_physics & clustering\_Spectral\_BERT \\ \hline
    computer\_security & clustering\_NMF\_TFIDF \\ \hline
    conceptual\_physics & clustering\_Spectral\_BERT \\ \hline
    econometrics & clustering\_NMF\_TFIDF \\ \hline
    electrical\_engineering & quality\_spelling\_error \\ \hline
    elementary\_mathematics & quality\_lexical\_diversity  \\ \hline
    formal\_logic & clustering\_Spectral\_BERT \\ \hline
    global\_facts & quality\_CPD \\ \hline
    high\_school\_biology & clustering\_Spectral\_MTEB \\ \hline
    high\_school\_chemistry & quality\_CPD \\ \hline
    high\_school\_computer\_science & quality\_spelling\_error \\ \hline
    high\_school\_european\_history & clustering\_Spectral\_BERT  \\ \hline
    high\_school\_geography & clustering\_NMF\_TFIDF \\ \hline
    high\_school\_macroeconomics & clustering\_NMF\_TFIDF  \\ \hline
    high\_school\_mathematics & clustering\_NMF\_TFIDF \\ \hline
    high\_school\_microeconomics & quality\_spelling\_error\\ \hline
    high\_school\_physics & quality\_spelling\_error\\ \hline
    high\_school\_psychology & random \\ \hline
    high\_school\_statistics & clustering\_NMF\_TFIDF \\ \hline
    high\_school\_us\_history & quality\_spelling\_error \\ \hline
    high\_school\_world\_history & clustering\_KMeans\_TFIDF  \\ \hline
    human\_aging & random \\ \hline
    human\_sexuality & clustering\_Spectral\_BERT \\ \hline
    international\_law & quality\_spelling\_error \\ \hline
    jurisprudence & clustering\_NMF\_TFIDF \\ \hline
    logical\_fallacies & random \\ \hline
    machine\_learning & quality\_spelling\_error  \\ \hline
    management & clustering\_Spectral\_BERT \\ \hline
    marketing & clustering\_KMeans\_TFIDF  \\ \hline
    medical\_genetics & quality\_lexical\_diversity \\ \hline
    miscellaneous & clustering\_NMF\_TFIDF  \\ \hline
    moral\_disputes & random \\ \hline
    moral\_scenarios & clustering\_NMF\_TFIDF \\ \hline
    nutrition & clustering\_Spectral\_BERT \\ \hline
    philosophy & quality\_spelling\_error \\ \hline
    prehistory & quality\_lexical\_diversity \\ \hline
    professional\_accounting & random \\ \hline
    professional\_law & clustering\_NMF\_TFIDF \\ \hline
    professional\_medicine & clustering\_Spectral\_MTEB  \\ \hline
    professional\_psychology & quality\_CPD \\ \hline
    public\_relations & clustering\_KMeans\_TFIDF \\ \hline
    security\_studies & clustering\_KMeans\_TFIDF  \\ \hline
    sociology & quality\_spelling\_error \\ \hline
    us\_foreign\_policy & clustering\_NMF\_TFIDF \\ \hline
    virology & clustering\_Spectral\_MTEB \\ \hline
    world\_religions & quality\_CPD \\ \hline
    \end{tabular}
    \label{tab:hendrycks_adaptive}
\end{table}

\subsection{SubLIME MI - Experiments}
SubLIME from Aesthetic and Clip Score leaderboard across various benchmarks are shown below. 
\begin{figure*}[h]
    \centering
    \includegraphics[width=\linewidth]{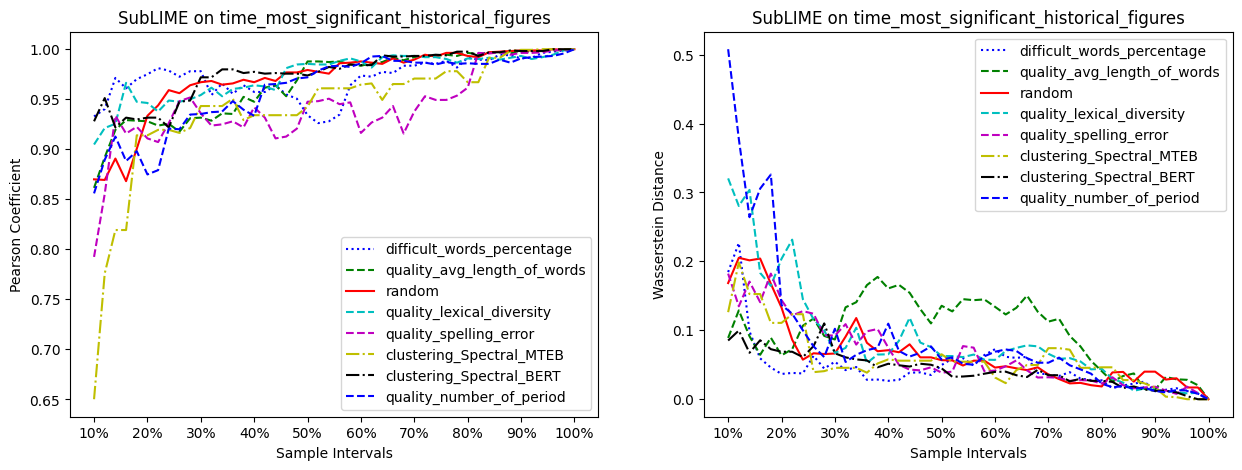}
    \caption{Rank and Score Preservation vs sampling ratio on most\_significant\_historical\_figures Benchmark on Aesthetics scores}
    \label{fig:aesthetic_rank_score}
\end{figure*}
\begin{figure*}[h]
    \centering
    \includegraphics[width=\linewidth]{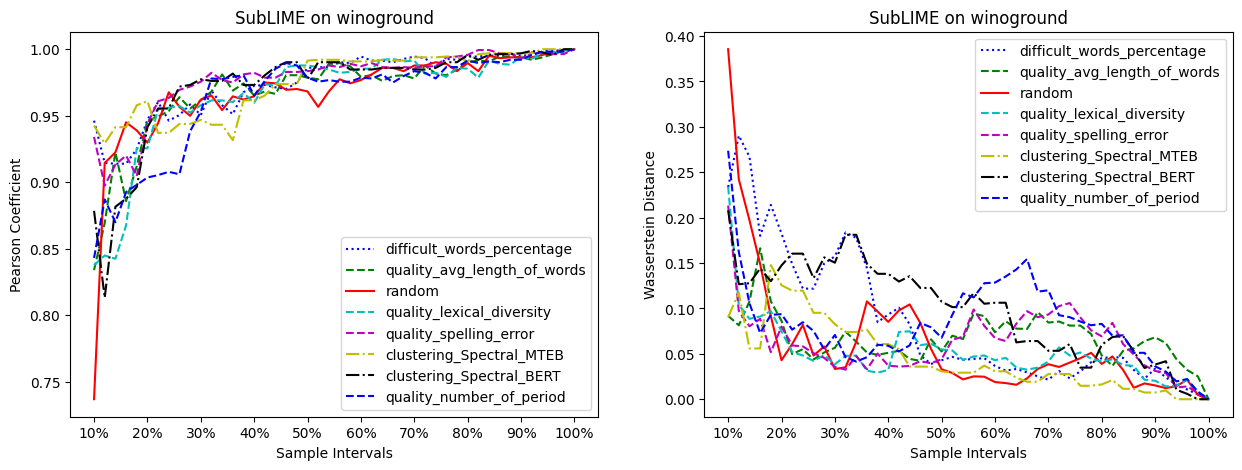}
    \caption{Rank and Score Preservation vs sampling ratio on winoground Benchmark on Aesthetics scores}
    \label{fig:aesthetic_winoground}
\end{figure*}
\begin{figure*}[h]
    \centering
    \includegraphics[width=\linewidth]{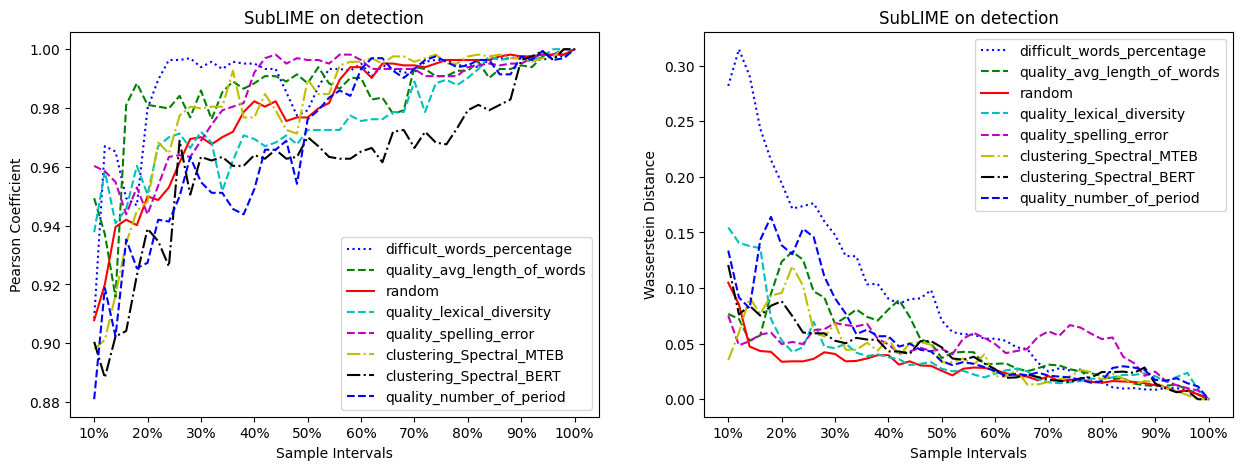}
    \caption{Rank and Score Preservation vs sampling ratio on detection Benchmark on Aesthetics scores}
    \label{fig:aesthetic_detection}
\end{figure*}
\begin{figure*}[h]
    \centering
    \includegraphics[width=\linewidth]{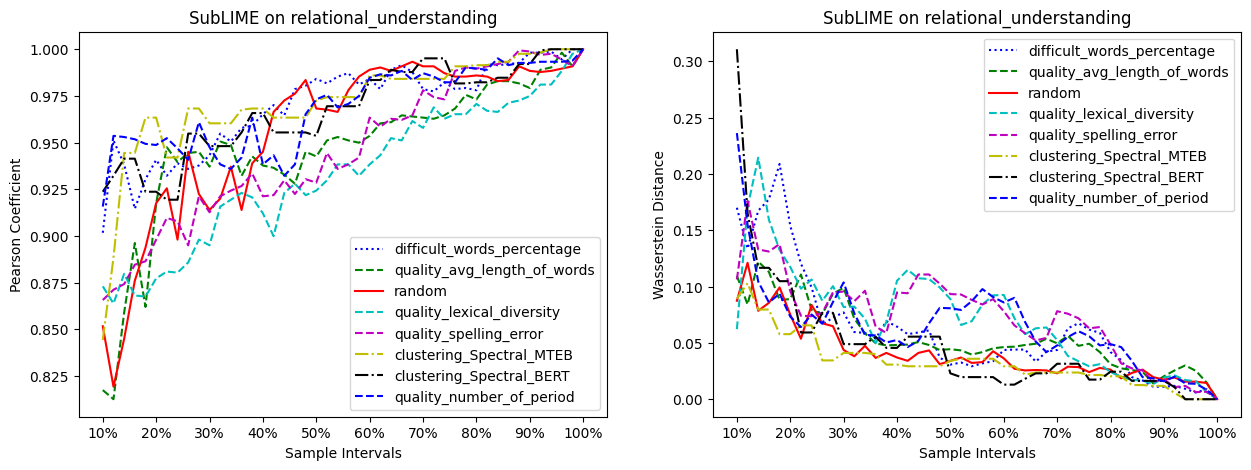}
    \caption{Rank and Score Preservation vs sampling ratio on Relational understanding Benchmark on Aesthetics scores}
    \label{fig:aesthetic_relational_understanding}
\end{figure*}

\begin{figure*}[h]
    \centering
    \includegraphics[width=\linewidth]{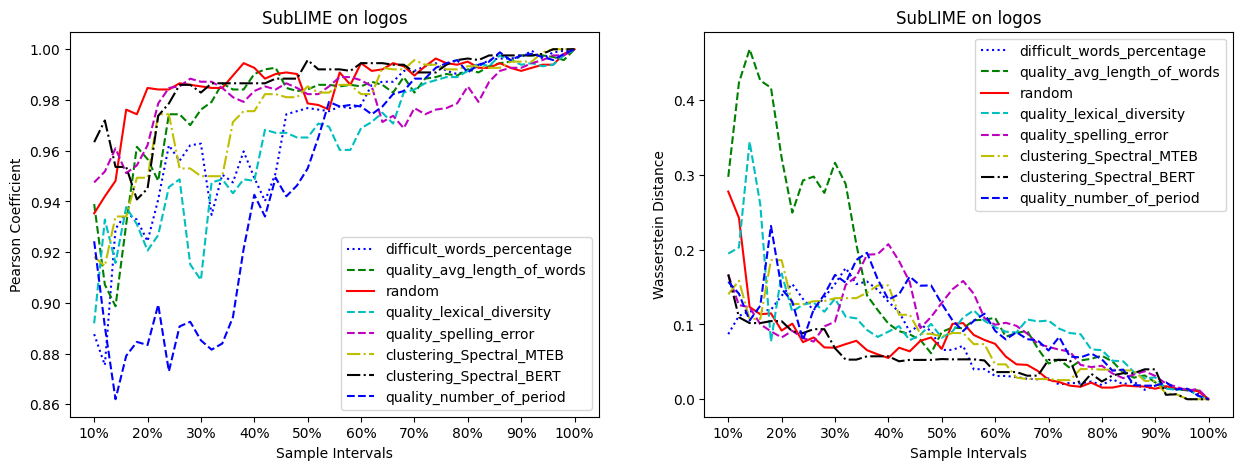}
    \caption{Rank and Score Preservation vs sampling ratio on Logos Benchmark on Aesthetics scores}
    \label{fig:aesthetic_logos}
\end{figure*}
\begin{figure*}[h]
    \centering
    \includegraphics[width=\linewidth]{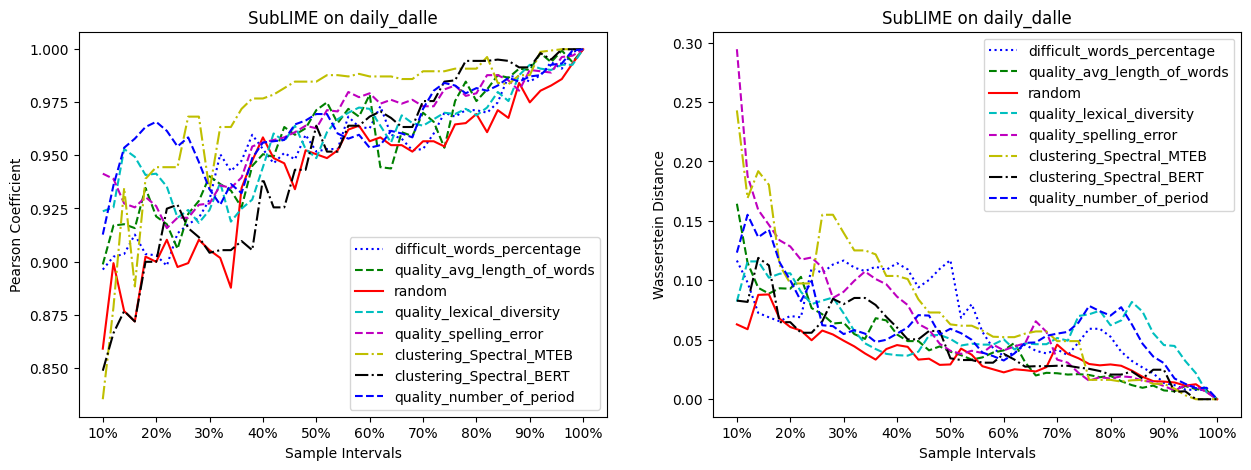}
    \caption{Rank and Score Preservation vs sampling ratio on Daily dalle Benchmark on Aesthetics scores}
    \label{fig:aesthetic_daily_dale}
\end{figure*}
\begin{figure*}[h]
    \centering
    \includegraphics[width=\linewidth]{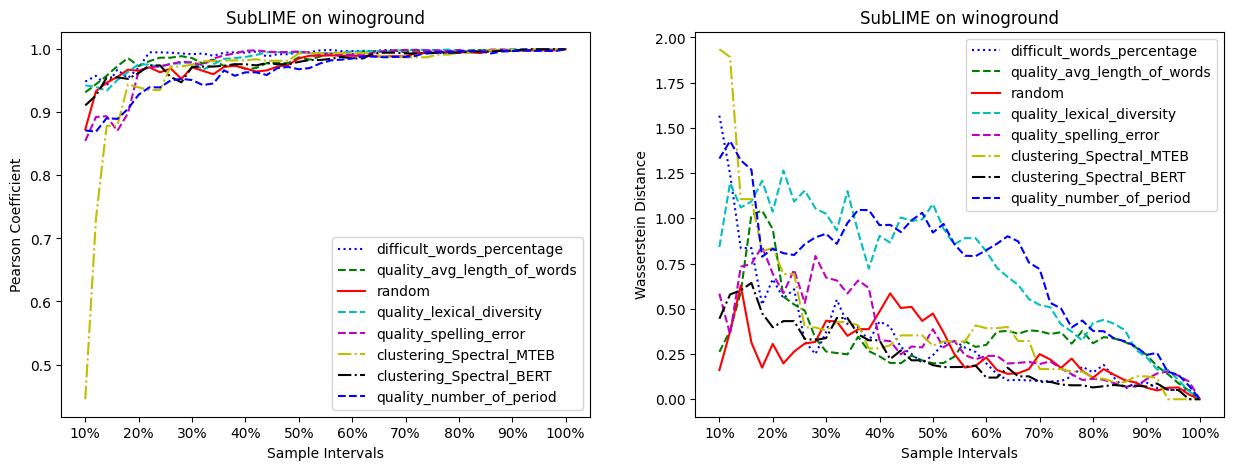}
    \caption{Rank and Score Preservation vs sampling ratio on Winoground Benchmark on Clip-scores}
    \label{fig:clip_winoground}
\end{figure*}
\begin{figure*}[h]
    \centering
    \includegraphics[width=\linewidth]{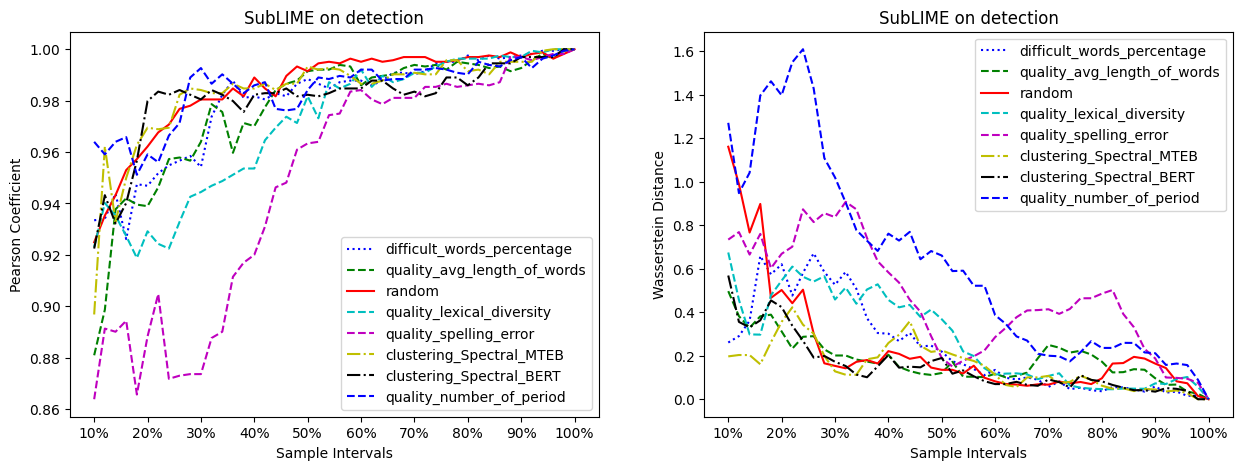}
    \caption{Rank and Score Preservation vs sampling ratio on Detection Benchmark on Clip-scores}
    \label{fig:clip_detection}
\end{figure*}
\begin{figure*}[h]
    \centering
    \includegraphics[width=\linewidth]{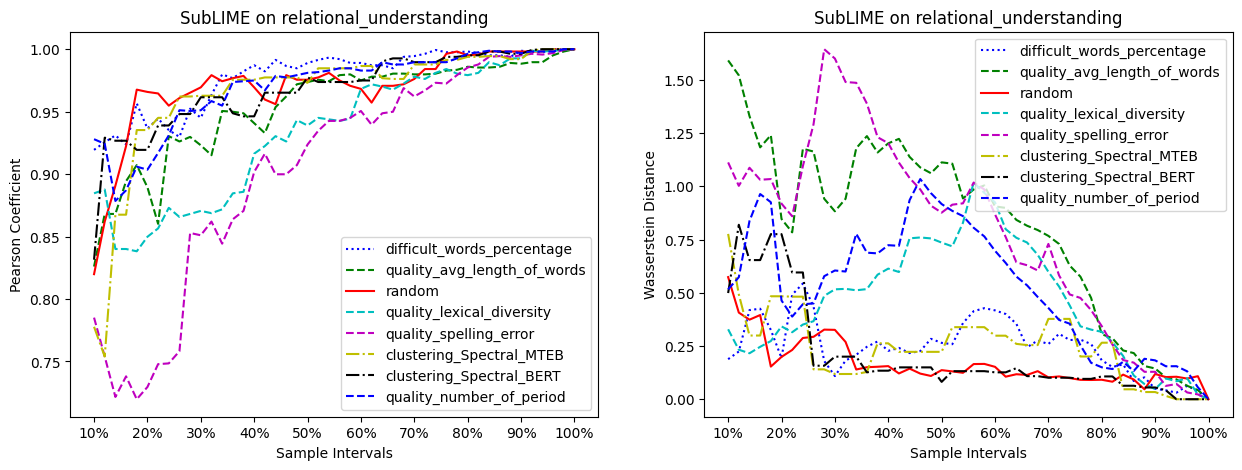}
    \caption{Rank and Score Preservation vs sampling ratio on Relational Understanding Benchmark on Clip-scores}
    \label{fig:clip_relational_understanding}
\end{figure*}
\begin{figure*}[h]
    \centering
    \includegraphics[width=\linewidth]{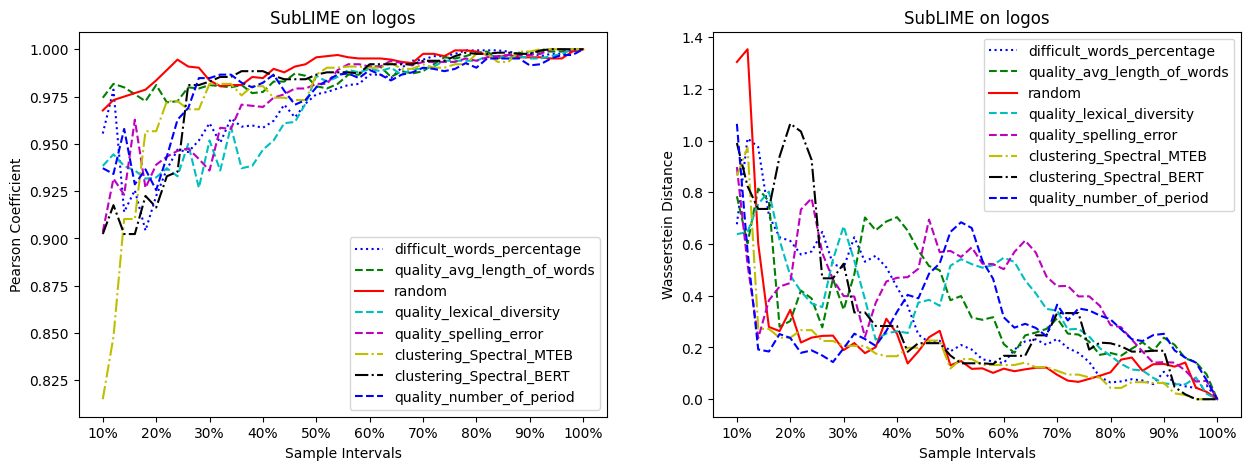}
    \caption{Rank and Score Preservation vs sampling ratio on Logos Benchmark on Clip-scores}
    \label{fig:clip_logos}
\end{figure*}
\begin{figure*}[h]
    \centering
    \includegraphics[width=\linewidth]{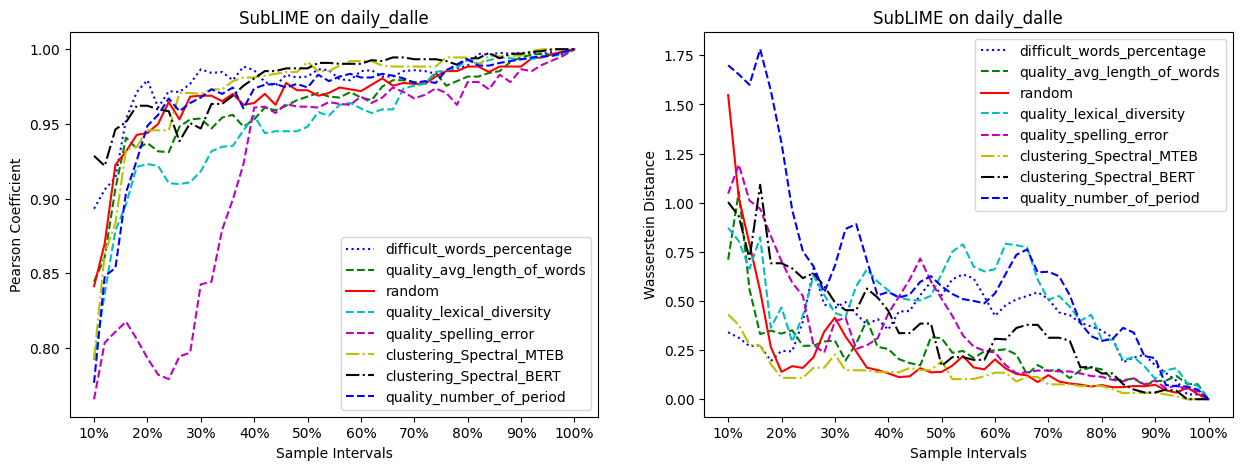}
    \caption{Rank and Score Preservation vs sampling ratio on Daile Dalle Benchmark on Clip-scores}
    \label{fig:clip_daile_dalle}
\end{figure*}
\begin{figure*}[h]
    \centering
    \includegraphics[width=\linewidth]{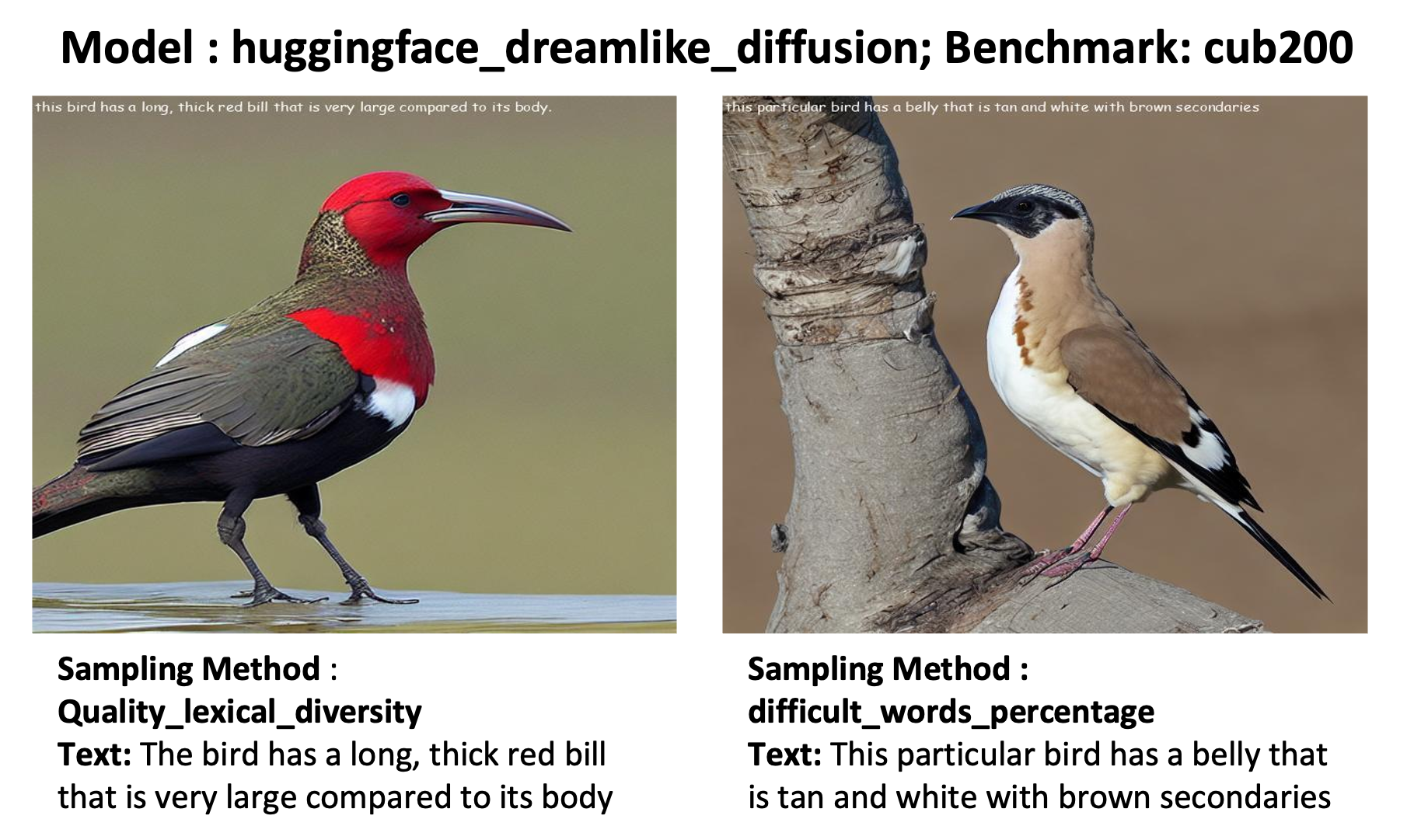}
    \caption{Generated Images from Cub200 Benchmark on Quality and Difficulty Sampling}
    \label{fig:cub200_gen_images}
\end{figure*}

\subsection{SubLIME MD - Experiments}
SubLIME on Individual Benchmarks Results are given below, this was performed before the cross benchmark evaluation. The Rank and Score preservation is performed on \textit{MBP} and \textit{HumanEval} using its result score. The redundancy is then remvoed with the help of GPT4.0 evaluator. 
\begin{figure*}[h]
    \centering
    \includegraphics[width=\linewidth]{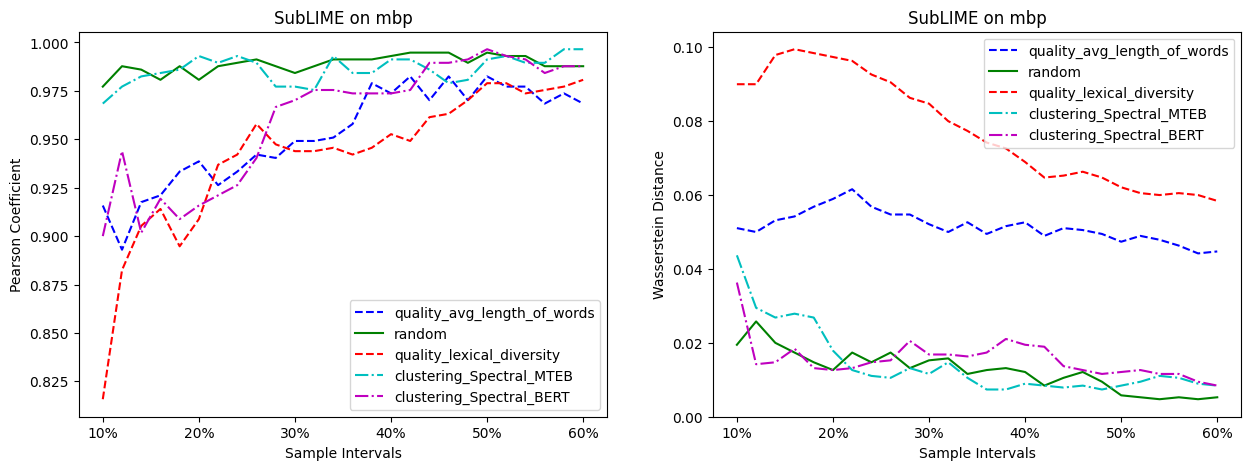}
    \caption{Rank and Score Preservation vs sampling ratio on MBP Benchmark: Before Redundancy Removal}
    \label{fig:mbp_bench}
\end{figure*}
\begin{figure*}[h]
    \centering
    \includegraphics[width=\linewidth]{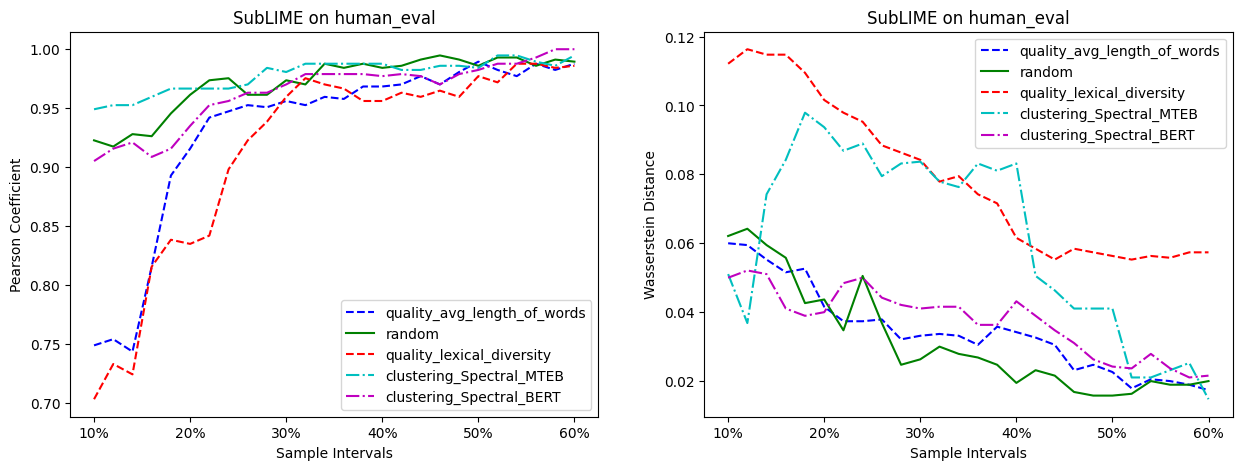}
    \caption{Rank and Score Preservation vs sampling ratio on HumanEval Benchmark: Before Redundancy Removal}
    \label{fig:humaneval_bench}
\end{figure*}
\begin{figure*}[h]
    \centering
    \includegraphics[width=\linewidth]{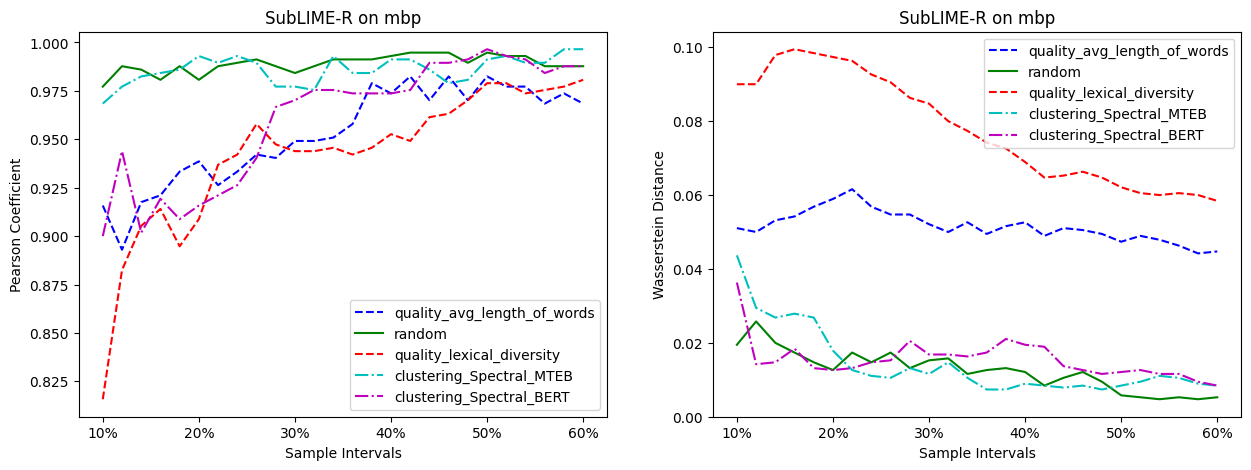}
    \caption{Rank and Score Preservation vs sampling ratio on MBP Benchmark: After Redundancy Removal}
    \label{fig:rmbp_bench}
\end{figure*}
\begin{figure*}[h]
    \centering
    \includegraphics[width=\linewidth]{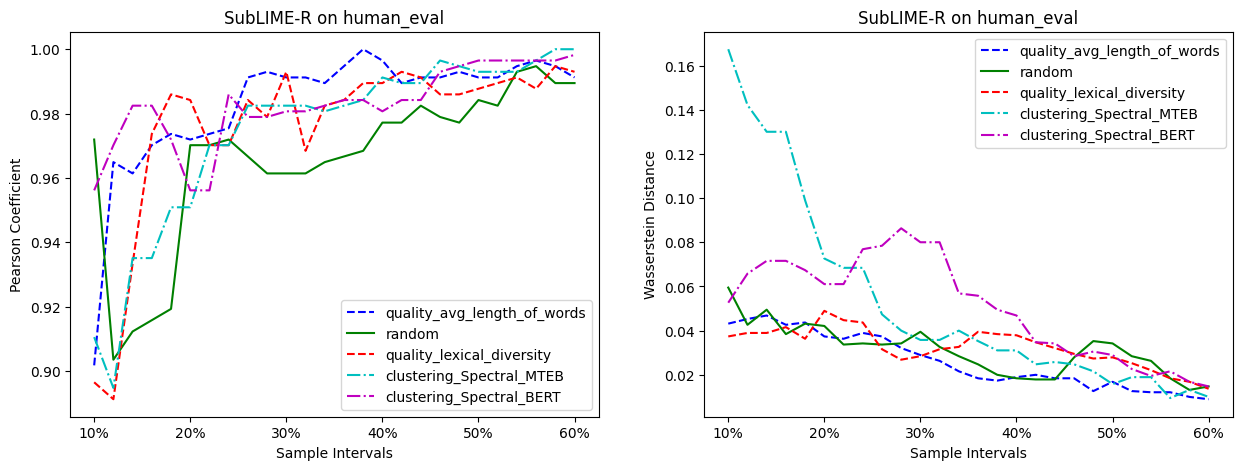}
    \caption{Rank and Score Preservation vs sampling ratio on HumanEval Benchmark: After Redundancy Removal}
    \label{fig:rhumaneval_bench}
\end{figure*}
\subsection{Discussion: Broader Applications of Adaptive Sampling}

\textbf{Tackling Unbalanced Benchmark}
Our analysis finds imbalances within certain benchmarks, i.e. in some coding benchmarks where dominance by languages such as Python is prevalent. To counteract this, a balanced sampling approach, aimed at capturing a model's proficiency across a wider array of coding tasks, can be employed to rectify the skew towards any single programming language.

\textbf{Enhancing Benchmark Fairness by Mitigating Bias}
Our adaptive sampling approach also could help address biases inherent in benchmarks, which can distort the evaluation outcomes. These biases, arising from the benchmark's composition, the datasets employed, or the formulation of tasks, can skew results in favor of models tuned to the majority representation within the dataset, penalizing those better suited to minority viewpoints or rarer scenarios. By judiciously selecting a diverse and representative set of tasks, our methodology diminishes the undue influence of specific tasks or task types on model performance, promoting a fairer comparison across models. 

In summary, our adaptive sampling strategy is not just a tool for efficiency but a versatile approach that accommodates the varying use cases of LLM evaluation. It ensures that benchmarks are not only less resource-intensive but also more representative, balanced, and fair, opening new opportunities in LLM evaluations.



\end{document}